\documentclass{llncs}
\usepackage{times}
\usepackage{color}
\usepackage{graphicx}
\usepackage[lined, ruled, boxed, commentsnumbered]{algorithm2e}
\usepackage{verbatim}
\usepackage{amsmath, amsfonts}
\usepackage{booktabs}
\usepackage{multirow}
\usepackage{floatrow}
\newfloatcommand{capbtabbox}{table}[][0.4\columnwidth]
\newcommand{\squishlist}{
 \begin{list}{$\bullet$}
  { \setlength{\itemsep}{0pt}
     \setlength{\parsep}{3pt}
     \setlength{\topsep}{3pt}
     \setlength{\partopsep}{0pt}
     \setlength{\leftmargin}{1.5em}
     \setlength{\labelwidth}{1em}
     \setlength{\labelsep}{0.5em} } }

\newcommand{\squishlisttwo}{
 \begin{list}{$\bullet$}
  { \setlength{\itemsep}{0pt}
     \setlength{\parsep}{0pt}
    \setlength{\topsep}{0pt}
    \setlength{\partopsep}{0pt}
    \setlength{\leftmargin}{2em}
    \setlength{\labelwidth}{1.5em}
    \setlength{\labelsep}{0.5em} } }

\newcommand{\squishlistthree}{
 \begin{enumerate}
  { \setlength{\itemsep}{0pt}
     \setlength{\parsep}{3pt}
     \setlength{\topsep}{3pt}
     \setlength{\partopsep}{0pt}
     \setlength{\leftmargin}{1.5em}
     \setlength{\labelwidth}{1em}
     \setlength{\labelsep}{0.5em} } }

\newcommand{\squishend}{\end{list}}
\newcommand{\squishendthree}{\end{enumerate}}

\begin{document}

\title{Credible Review Detection with Limited Information using Consistency Features}

\author{Subhabrata Mukherjee, Sourav Dutta, \and Gerhard Weikum}
\institute{Max Planck Institute for Informatics, Germany \\
\email{\{smukherjee,sdutta,weikum\}@mpi-inf.mpg.de}}

\maketitle

\begin{abstract}
Online reviews provide viewpoints on the strengths and shortcomings of products/services, influencing potential customers' purchasing decisions. 
However, the proliferation of \textit{non-credible} reviews --- either fake (promoting/ demoting an item)
, incompetent (involving irrelevant aspects), or
biased --- entails the problem of identifying {\em credible} reviews.
Prior works involve classifiers harnessing rich information about items/users --- 
which might not be readily available in several domains --- 
that 
provide only limited 
interpretability
as to why a review is deemed non-credible.

This paper presents a novel approach to address the above issues.
We utilize latent topic models leveraging review texts, item ratings, and timestamps to derive {\em consistency features}
without relying on item/user histories, unavailable for ``long-tail'' items/users.
We develop models, for computing review credibility scores to provide interpretable evidence for non-credible reviews, that are also transferable to other domains 
--- addressing the scarcity of labeled data.
Experiments on real-world
datasets demonstrate improvements 
over state-of-the-art baselines.
\end{abstract}

\section{Introduction}
\label{sec:intro}

{\bf Motivation:} 
Online reviews about hotels, restaurants, consumer goods, movies, books, drugs, etc. are an invaluable resource for Internet users, providing a wealth of related information for potential 
customers. Unfortunately, corresponding forums such as TripAdvisor, Yelp, Amazon, and others are being increasingly game to manipulative and deceptive reviews: fake (to promote or demote some item), 
incompetent (rating an item based on irrelevant aspects), or biased (giving a distorted and inconsistent view of the item). For example, recent studies depict that $20\%$ of Yelp reviews might be 
fake and Yelp internally rejects $16\%$ of user submissions~\cite{Luca} as ``not-recommended''; with similar figures reported for reviews on Amazon.

Starting with the work of \cite{Liu2008}, research efforts have been undertaken to automatically detect non-credible reviews. In parallel, industry (e.g., stakeholders such as Yelp) has 
developed its own standards\footnote{\url{officialblog.yelp.com/2009/10/why-yelp-has-a-review-filter.html}} to filter out ``illegitimate'' reviews. Although details are not disclosed, studies suggest that these filters tend to be fairly 
crude~\cite{Liu2013a}; for instance, exploiting user activity like the number of reviews posted, and treating users whose ratings show high deviation from the mean/majority ratings as suspicious. Such a policy seems to 
over-emphasize trusted long-term contributors and suppress outlier opinions off the mainstream. Moreover, these filters also employ several aggregated metadata, and are thus hardly viable for 
new items that initially have very few reviews --- often by not so active users or newcomers in the community.

\noindent {\bf State of the Art:}
Research on this topic has cast the problem of review credibility into a binary classification task: a review is either credible or deceptive.
To this end, supervised and semi-supervised methods have been developed that largely rely on features about users and their activities as well as statistics about item ratings. 
Most techniques also consider spatio-temporal patterns of user activities like 
IP addresses or user locations (e.g., \cite{Liu2014,Li2015}), 
burstiness of posts on an item or an item group (e.g., \cite{Fei2013}), 
and further 
correlation measures across users and items (e.g., \cite{mukherjee2014peopleondrugs}). 
However, the classifiers built this way are mostly geared for popular items, and the meta-information about user histories and activity correlations are not always available. For example, 
someone interested in opinions on a new art film or a ``long-tail'' bed-and-breakfast in a rarely visited town, is not helped at all by the above methods. 
%
Several existing works~\cite{Mihalcea2009,Ott2011,Ott2013} consider the textual content of user reviews for tackling opinion spam by using word-level unigrams or bigrams as features, 
along with specific lexicons (e.g., LIWC~\cite{liwc} psycholinguistic lexicon, WordNet Affect~\cite{wordnetaffect}), to learn latent topic models and classifiers (e.g., \cite{Ott2013a}).
Although these methods achieve high classification accuracy for various gold-standard datasets, they do not provide any interpretable evidence as to why a certain review is classified as 
non-credible.

\noindent {\bf Problem Statement:}
This paper focuses on detecting credible reviews {\em with limited information}, namely, in the absence of rich data about user histories, community-wide correlations, and for ``long-tail'' items. 
In the extreme case, we are provided with only the review texts and ratings for an item. Our goal is then to compute a {\em credibility score} for the reviews and to provide possibly 
{\em interpretable evidence} for explaining why certain reviews have been categorized as non-credible.

\noindent {\bf Approach:}
Our proposed method to this end is to learn a model based on {\em latent topic models} and combining them with limited metadata to provide a novel notion of {\em consistency features} 
characterizing each review. We use the LDA-based Joint Sentiment Topic model (JST)~\cite{Lin2009} to cast the user review texts into a number of informative facets. We do this per-item, 
aggregating the text among all reviews for the same item, and also per-review. This allows us to identify, score, and highlight inconsistencies that may appear between a review and 
the community's overall characterization of an item. We perform this for the item as a whole, and also for each of the latent facets separately.
Additionally, we learn inconsistencies such as discrepancy between the contents of a review and its rating, and temporal ``bursts'' --- where a number of reviews are written in a short span 
of time targeting an item. We propose five kinds of inconsistencies that form the key assets of our credibility scoring model, fed into a Support Vector Machine for classification, or for ordinal ranking.

\noindent {\bf Contribution:} In summary, our contributions are summarized as:
\squishlist
 \item {\em Model:} We develop a novel {\em consistency model} for credibility analysis of reviews that works with limited information, with particular attention to ``long-tail'' items, and offers interpretable evidence for reviews classified as non-credible.
 \item {\em Tasks:} We investigate how credibility scores affect the overall ranking of items. To address the scarcity of labeled training data, we transfer the learned model from Yelp to Amazon to rank top-selling items based on (classified) {\em credible} user reviews. In the presence of proxy labels for item ``goodness'' (e.g., item sales rank), we develop a better ranking model for domain adaptation.
 \item {\em Experiments:} We perform extensive experiments in TripAdvisor, Yelp, and Amazon to demonstrate the viability of our method and its advantages over state-of-the-art baselines in dealing with ``long-tail'' items and providing interpretable evidence.
\squishend

\section{Related Work}

Previous works in fake review and opinion spam detection primarily focused on two different aspects of the problem: 

\noindent {\bf Linguistic Analysis}~\cite{Mihalcea2009,Ott2011,Ott2013} -- This approach exploits the distributional difference in the wordings of authentic and manually-created 
fake reviews using word-level features. However, artificially created fake review datasets for the studied tasks give away explicit features not dominant in 
real-world data. This was confirmed by a study on Yelp filtered reviews~\cite{Liu2013a}, where the $n$-gram features performed poorly despite their outstanding performance 
on the Amazon Mechanical Turk generated fake review dataset. Additionally, linguistic features such as {\em text sentiment}~\cite{Yoo2009}, {\em readability score} 
(e.g., Automated readability index (ARI), Flesch reading ease, etc.)~\cite{Hu2012}, {\em textual coherence}~\cite{Mihalcea2009}, and rules based on 
{\em Probabilistic Context Free Grammar} (PCFG)~\cite{Feng2012} have been studied.

\noindent {\bf Rating and Activity Analysis} -- In the absence of proper ground-truth data, prior works make simplistic assumptions, e.g., duplicates and near-duplicates are 
fake, and make use of {\em extensive} background information like brand name, item description, user history, IP addresses and location, etc.~\cite{Liu2007,Liu2008,Liu2010,Liu2011,Liu2012,Liu2013,Liu2013a,Liu2014,Rahman2015}. 
Thereafter, regression models trained on all these features are used to classify reviews as credible or deceptive. Some of these works also use crude or ad-hoc language 
features like content similarity, presence of literals, numerals, and capitalization.
In contrast to these works, our approach uses limited information about users and items catering to a broad domain of applications. We harvest several consistency features from user 
rating and review text that give some interpretation as to why a review should be deemed non-credible.

\noindent {\bf Learning to Rank} -- Supervised models have also been developed to rank items from constructed item feature vectors~\cite{liu2009}. Such techniques optimize measures like Discounted Cumulative Gain, Kendall-Tau, and Reciprocal Rank to generate item ranking similar to the training data based on the feature vectors. We use one such technique, and show its performance can be improved by removing non-credible item reviews.

\section{Review Credibility Analysis}

\subsection{Language Model}
\label{subsec:lang}

Previous works~\cite{Ott2011,Ott2013,Mihalcea2009,Chen2015} in linguistic analysis explore distributional difference in the wordings between deceptive and authentic reviews. In general, authentic reviews tend to have more {\em sensorial and concrete language} than deceptive reviews, with higher usage of nouns, adjectives, prepositions, determiners, and coordinating conjunctions; whereas deceptive reviews were shown to use more verbs, adverbs, and superlatives manifested in exaggeration for imaginary writing. \cite{Ott2011,Ott2013} found that authentic hotel reviews are more specific about spatial configurations (small room, low ceiling, etc.) and aspects like location, amenities and cost; whereas deceptive reviews focus on aspects external to the item being reviewed (like traffic jam, children, business, and vacation). Extreme opinions were also found to be dominant in deceptive reviews to assert stances, whereas authentic reviews have a more balanced view analyzing the item on several aspects. We implicitly 
exploit these features in the latent facet model (discussed in the next section) to find the reviewer opinion on important facets of the item under consideration, and the overall rating distribution obtained from facet level opinions.

In order to explicitly capture such distributional difference in the language of credible and non-credible reviews at word-level, we use unigram and bigram language features that have been shown to outperform other fine-grained linguistic features using psycholinguistic features (e.g., LIWC lexicon) and Part-of-Speech tags~\cite{Ott2011}. We also experimented with WordNet Affect to capture fine-grained emotional dimensions (like anger, hatred, and confidence), which, however, were seen not to perform well. In general, the bigram features capture context-dependent information to some extent, and together with simple unigram features performed the best. We also observed that the presence or absence of words, mattered more than their frequency for credibility analysis. In our model, all the features were length normalized, retaining punctuations (like `!') and capitalization as non-credible reviews manifesting exaggeration tend to over-use the latter features (e.g., ``the hotel was AWESOME !!!'').

\noindent {\bf Feature vector construction}: Consider a vocabulary $V$ of unique unigrams and bigrams in the corpus (after removing stop words). For each token type $f_i \in V$ 
and each review $d_j$, we compute the presence/absence of words, $w_{ij}$, of type $f_i$ occurring in $d_j$, thus constructing a feature vector $F^L(d_j) = \langle w_{ij}=I({w_{ij} = f_i}) ~ / ~length(d_j) \rangle, \forall i$, with $I(.)$ denoting an indicator function (notations used are presented in Table~\ref{tab:notation}).

\subsection{Facet Model}

Given review snippets like ``the hotel offers free wi-fi'', we now aim to find the different facets present in the reviews along with their corresponding sentiment polarities. Since the aim of this work is to present a model requiring limited prior information, we extract the {\em latent} facets from the review text, without the help of any explicit facet or seed words. The ideal machinery should map ``wi-fi'' to a latent facet cluster like ``network, Internet, computer, access, ...''. We also want to extract the sentiment expressed in the review about the facet. Interestingly, although ``free'' does not 
have a polarity of its own, in the above example ``free'' in conjunction with ``wi-fi'' expresses a positive sentiment of a service being offered without charge. The hope is that although ``free'' does not have an individual polarity, it appears in the neighborhood of words that have known polarities (from lexicons). This helps in the joint discovery of facets and sentiment labels, as ``free wi-fi'' and ``internet without extra charge'' should ideally map to the same facet cluster with similar polarities using their co-occurrence with similar words with positive polarities. In this work, we use the Joint Sentiment Topic Model approach (JST)~\cite{Lin2009} to jointly discover the latent facets along with their expressed polarities.


Consider a set of reviews $\langle D \rangle$ written by users $\langle U \rangle$ on a set of items $\langle I \rangle$, with $r_d \in R$ being the rating assigned to review $d \in D$. Each review document $d$ consists of a sequence of words $N_d$ denoted by $\{w_1, w_2, ... w_{N_d}\}$, and each word is drawn from a vocabulary $V$ indexed by $1,2,..V$. Consider a set of facet assignments $z=\{z_1, z_2, ... z_K\}$ and sentiment label assignments $l = \{l_1, l_2, ... l_L\}$ for $d$, where each $z_i$ can be from a set of $K$ possible facets, and each label $l_i$ is from a set of $L$ possible sentiment labels. 

JST adds a layer of sentiment in addition to the topics as in standard LDA~\cite{Blei2003}. It assumes each document $d$ to be associated with a multinomial distribution $\theta_d$ over facets $z$ and sentiment labels $l$ with a symmetric Dirichlet prior $\alpha$. $\theta_d(z,l)$ denotes the probability of occurrence of facet $z$ with polarity $l$ in document $d$. Topics have a multinomial distribution $\phi_{z,l}$ over words drawn from a vocabulary $V$ with a symmetric Dirichlet prior $\beta$. $\phi_{z,l}(w)$ denotes the probability of the word $w$ belonging to the facet $z$ with polarity $l$. In the generative process, a sentiment label $l$ is first chosen from a document-specific rating distribution $\pi_d$ with a symmetric Dirichlet prior $\gamma$ . 
Thereafter, one chooses a facet $z$ from $\theta_d$ conditioned on $l$, and subsequently a word $w$ from $\phi$ conditioned on $z$ and $l$. Exact inference is not possible due to 
intractable coupling between $\Theta$ and $\Phi$, and thus we use Collapsed Gibbs Sampling for approximate inference.

Let $n(d, z, l, w)$ denote the count of the word $w$ occurring in document $d$ belonging to the facet $z$ with polarity $l$. The conditional distribution for the latent variable $z$ (with components $z_1$ to $z_K$) and $l$ (with components $l_1$ to $l_L$) is given by:
%
\begin{equation}
\label{eq.3}
\begin{aligned}
 P(z_i=k, l_i=j| &w_i=w, z_{-i}, l_{-i}, w_{-i}) \propto\\
  \frac{n(d, k, j, .) + \alpha}{\sum_{k}n(d, k, j, .) + K \alpha} &\times \frac{n(., k, j, w) + \beta}{\sum_{w}n(., k, j, w) + V \beta} \times \frac{n(d,.,j,.) + \gamma}{\sum_j n(d, ., j,.) + L \gamma}
 \end{aligned}
\end{equation}
In the above equation, the operator $(.)$ in the count indicates marginalization, i.e.,  summing up the counts over all values for the corresponding position in $n(d,z,l,w)$, and 
the subscript $-i$ denotes the value of a variable excluding the data at the $i^{th}$ position.

\subsection{Consistency Features}
\label{subsec:cons}

We extract the following features from the latent facet model enabling us to detect {\em inconsistencies} in reviews and ratings of items for credibility analysis.

\noindent 1. {\bf User Review -- Facet Description:} The facet-label distribution of different items differ; for some items, certain facets (with their polarities) are more important than 
other dimensions. For instance, the ``battery life'' and ``ease of use'' for consumer electronics are more important than ``color''; for hotels, certain services are available for free (e.g., wi-fi) 
which may be charged elsewhere. Similarly, user reviews involving less relevant facets of the item under discussion, e.g., downrating hotels for ``not allowing pets'' should also be detected.

Given a review $d(i)$ on an item $i \in I$ with a sequence of words $\{w\}$ and previously learned $\Phi$, its facet label distribution $\Phi^{'}_d(i)$ with dimension $K \times L$ is given by:
\begin{equation}
 \phi^{'}_{k,l} = \sum_{w: l^{*} = argmax_{l}\ \phi_{k,l}(w)} \phi_{k, l^{*}}(w)
\end{equation}
For each word $w$ and each latent facet dimension $k$, we consider the sentiment label $l^{*}$ that maximizes the facet-label-word distribution $\phi_{k,l}(w)$, and aggregate this over all the words.
This facet-label distribution of the review $\Phi^{'}_d(i)$ of dimension $K \times L$ is used as a feature vector to a classifier to figure out the importance of the different latent dimensions that 
also captures {\em domain-specific} facet-label importance.

\noindent 2. {\bf User Review --- Rating:} The user-assigned rating corresponding to the review should be consistent to her opinion expressed in the review text. For example, the user is unlikely to give 
an average rating to an item when she expresses a positive opinion about all the important facets of the item. 
The inferred rating distribution $\pi^{'}_d $ (with dimension $L$) of a review $d$ consisting of a sequence of words $\{w\}$ and learned $\Phi$ is computed as:
\begin{equation}
 \pi^{'}_l = \sum_{w,k:\{k^{*}, l^{*}\} = argmax_{k,l}\ \phi_{k,l}(w)} \phi_{k^{*}, l^{*}}(w)
\end{equation}
For each word, we consider the facet and label that jointly maximizes the facet-label-word distribution, and aggregate over all the words and facets. The absolute deviation (of dimension $L$) between the user-assigned 
rating $\pi_d$, and estimated rating $\pi^{'}_d$ from user text is taken as a component in the overall feature vector.

\noindent 3. {\bf User Rating:} Previous works~\cite{Ott2011,Sun2013,Hu2012} on opinion spam found that fake reviews tend to have overtly positive or overtly negative opinions. Therefore, we also 
use $\pi^{'}_d$ as a component of the overall feature vector to detect cues from such extreme ratings.

\noindent 4. {\bf Temporal Burst:} This is typically observed in {\em group spamming}, where a number of reviews are posted targeting an item in a short span of time. Consider a set of reviews 
$\{d_j\}$ at timepoints $\{t_j\}$ posted for a {\em specific} item. The temporal burstiness of review $d_i$ for the given item is given by $\big( \sum_{j, j\neq i} \frac{1}{1 + e^{t_i-t_j}} \big)$. 
Here, exponential decay is used to weigh the temporal proximity of reviews to capture the burst.

\noindent 5. {\bf User Review -- Item Description:} In general, the description of the facets outlined in a user review about an item should not differ much from that of the majority. 
For example, if majority says the ``hotel offers free wi-fi'', and the user review says ``internet is charged'' --- this presents a possible inconsistency. For the facet model this corresponds to word 
clusters having the same facet label but different sentiment labels. During experiments, however, we find this feature to play a weak role in the presence of other inconsistency features.

We aggregate the {\em per-review} facet distribution $\phi^{'}_{k,l}$ over all the reviews $d(i)$ on the item $i$ to obtain the facet-label distribution $\Phi^{''}(i)$ of the item. We use the Jensen-Shannon divergence, a symmetric and smoothed version of the Kullback-Leibler divergence as a feature. This depicts how much the facet-label distribution in the given review diverges from the general opinion of other people about the item.
\begin{equation}
 JSD(\Phi^{'}_d(i)\ ||\ \Phi^{''}(i)) = \frac{1}{2} (D(\Phi^{'}_d(i)\ ||\ M) + D(\Phi^{''}(i)\ ||\ M))
\end{equation}
where, $M = \frac{1}{2}(\Phi^{'}_d(i)+\Phi^{''}(i))$, and $D$ represents Kullback-Leibler divergence.

\noindent{\bf Feature vector construction}: For each review $d_j$, all the above {\em consistency features} are computed, and a facet feature vector $\langle F^T(d_j) \rangle$ of dimension 
$2 + K \times L + 2L$ is created for subsequent processing.

\subsection{Behavioral Model}
\label{subsec:user}

Earlier works~\cite{Liu2007,Liu2008,Liu2010} on review spam show that user-dependent models detecting user-preferences and biases perform well in credibility analysis. However, such information 
is not always available, especially for newcomers, and not so active users in the community.
Besides, ~\cite{Liu2012,Liu2013} show that spammers tend to open multiple fake accounts to write reviews for malicious activities --- using each of those accounts sparsely to avoid detection. 
Therefore, instead of relying on extensive user history, we use simple proxies for user activity that are easier to aggregate from the community:
\squishlistthree
\item {\bf User Posts:} number of posts written by the user in the community.
\item {\bf Review Length:} length of the reviews --- longer reviews tend to frequently go off-topic with high emotional digression.
\item {\bf User Rating Behavior:} absolute deviation of the review rating from the mean and median rating of the user to other items, as well as the first three moments of the user rating distribution 
--- capturing the scenario where the user has a {\em typical rating behavior} across all items.
\item {\bf Item Rating Pattern:} absolute deviation of the item rating from the mean and median rating obtained from other users captures the extent to which the user disagrees with other users about the item quality; the first three moments of the item rating distribution captures the general item rating pattern.
\item {\bf User Friends:} number of friends of the user.
\item {\bf User Check-in:} if the user checked-in the hotel --- first hand experience of the user adds to the review credibility.
\item {\bf Elite:} elite status of the user in the community.
\item {\bf Review helpfulness:} number of helpfulness votes received by the user post --- captures the quality of user postings.
\squishendthree
Note that user rating behavior and item rating pattern are also captured {\em implicitly} using the consistency features in the latent facet model. 


Since our aim is to detect credible reviews in the case of limited information, we further split the above activity or behavioral features into two components: (a) $Activity^{-}$ using features $[1-4]$ 
that can be straightforward obtained from the tuple $\langle userId, itemId, review, rating \rangle$ and are easily available even for ``long-tail'' items and newcomers; and (b) $Activity^{+}$ using all 
the listed features. However the latter requires additional information (features $[5-8]$) that might not always be available, or takes long time to aggregate for new items/users.

\noindent{\bf Feature vector construction}: For each review $d_j$ by user $u_k$, we construct a behavioral feature vector $\langle F^B(d_j) \rangle$ using the above features.

\subsection{Application Oriented Tasks}

\noindent {\bf Credible Review Classification:}
In the first task, we {\em classify} reviews as {\em credible} or not. For each review $d_j$ by user $u_k$, we construct the joint feature vector $F(d_j) = F^L(d_j) \cup F^T(d_j) \cup F^B(d_j)$, and 
use Support Vector Machines (SVM)~\cite{Cortes1995} for classification of the reviews. SVM maps the examples (using Kernels) to a high dimensional space, and constructs a hyperplane to 
separate the two categories of examples. Although there can be an infinite number of such hyperplanes possible, SVM constructs the one with the largest functional margin given by the distance of the 
nearest point to the hyperplane on each side of it. New points are mapped to the same space and classified to a category based on which side of the hyperplane it lies. We use a linear 
kernel 
which has been shown to perform the best for text classification tasks. We use the $L_2$ regularized $L_2$ loss SVM with dual formulation from the LibLinear package (csie.ntu.edu.tw/cjlin/liblinear)~\cite{Fan2008} with other default parameters. We report classification accuracy 
with $10$-fold cross-validation on ground-truth from TripAdvisor and Yelp.

\noindent {\bf Item Ranking:}
Due to the scarcity of ground-truth data pertaining to review credibility, a more suitable way to evaluate our model is to examine the {\em effect} of non-credible reviews on the relative 
{\em ranking} of items in the community. For instance, in case of popular items with large number of reviews, even if a fraction of it were non-credible, its effect would not be so severe 
as would be on ``long-tail'' items with fewer reviews. 

A simple way to find the ``goodness'' of an item is to aggregate ratings of all reviews -- using which we also obtain a ranking of items. We use our model to filter 
out non-credible reviews, aggregate ratings of credible reviews, and re-compute the item ranks.

\noindent {\em \bf Evaluation Measures} --
We use the {\em Kendall-Tau Rank Correlation Co-efficient} ($\tau$) to find effectiveness of the rankings, against a {\em reference ranking} --- for instance, the {\em sales rank} of items 
in Amazon.
$\tau$ measures the number of concordant and discordant pairs, to find whether the ranks of two elements agree or not based on their scores, out of the total number of combinations possible. 
Given a set of observations $\{x, y\}$, any pair of observations $(x_i, y_i)$ and $(x_j, y_j)$, where $i \not= j$, are said to be {\em concordant} if either $x_i > x_j$ and $y_i > y_j$, or 
$x_i < x_j$ and $y_i < y_j$, and {\em discordant} otherwise. If $x_i=x_j$ or $y_i=y_j$, the ranks are tied --- neither discordant, nor concordant.

We use the {\em Kendall-Tau-B} measure ($\tau_b$) which allows for rank adjustment. Consider $n_c$, $n_d$, $t_x$, and $t_y$ to be the number of concordant, discordant, tied pairs on $x$, and tied 
pairs on $y$ respectively, whereby Kendall-Tau-B is given by: $\frac{n_c-n_d}{\sqrt{(n_c + n_d + t_x)(n_c + n_d + t_y)}}$. 

However, this is a conservative estimate as multiple items --- typically the top-selling ones in Amazon --- have the same rating (say, $5$). Therefore, we use a second estimate 
(say, {\em Kendall-Tau-M} ($\tau_m$)) which considers non-zero tied ranks to be concordant. Note that, an item can have a zero-rank if all of its reviews are classified as non-credible. A high positive (or, negative) value of Kendall-Tau indicates the two series are positively (or, negatively) correlated; whereas a value close to zero indicates they are independent. 

\noindent {\em \bf Domain Transfer from Yelp to Amazon} -- 
A typical issue in credibility analysis task is the scarcity of labeled training data. In the first task, we use labels from the Yelp Spam Filter (considered to be the industry standard) to train our model. However, such ground-truth labels are not available in Amazon. Although, in principle, we can train a model $M_\text{Yelp}$ on Yelp, and use it to filter out non-credible reviews in Amazon.

Transferring the learned model from Yelp to Amazon (or other domains) entails using the learned weights of {\em features} in Yelp that are analogous to the ones in Amazon. However, this process 
encounters the following issues:
\squishlisttwo
\item Facet distribution of Yelp (food and restaurants) is different from that of Amazon (products such as software, and consumer electronics). Therefore, the facet-label distribution and the 
corresponding learned feature weights from Yelp cannot be directly used, as the latent dimensions are different. 
\item Additionally, specific metadata like check-in, user-friends, and elite-status are missing in Amazon.
\squishend
However, the learned weights for the following features can still be directly used:
\squishlisttwo
\item Certain unigrams and bigrams, especially those depicting opinion, that occur in both domains.
\item Behavioral features like user and item rating patterns, review count and length, and usefulness votes.
\item Deviation features derived from {\em Amazon-specific} facet-label distribution that is obtained using the JST model on Amazon corpus:
\squishlist
\item Deviation (with dimension $L$) of the user assigned rating from that inferred from review content.
\item Distribution (with dimension $L$) of positive and negative sentiment as expressed in the review.
\item Divergence, as a unary feature, of the facet-label distribution in the review from the aggregated distribution over other reviews on a given item.
\item Burstiness, as a unary feature, of the review.
\squishend
\squishend

Using the above components, that are common to both Yelp and Amazon, we {\em first} re-train the model $M_{\text{Yelp}}$ from Yelp to remove the non-contributing features for Amazon. 

Now, a direct transfer of the model weights from Yelp to Amazon assumes the distribution of credible to non-credible reviews, and corresponding feature importance, to be the same in both domains --- 
which is not necessarily true. In order to boost certain features to better identify non-credible reviews in Amazon, we tune the {\em soft margin parameter} $C$ in the SVM. We use 
{\em C-SVM}~\cite{chen04}, with slack variables, that optimizes:\\
$min_{\vec{w}, b, \xi_i \ge 0} \frac{1}{2}\vec{w}^T\vec{w}+C^{+}\sum_{y_i=+1}\xi_i+C^{-}\sum_{y_i=-1}\xi_i$ \\
subject to $\forall \{(\vec{x_i}, y_i)\}, y_i(\vec{w}^{T}\vec{x}_i + b) \ge 1- \xi_i$.

$C^{+}$ and $C^{-}$ are regularization parameters for positive and negative class (credible and deceptive), respectively. The parameters $\{C\}$ provide a trade off as to how wide the margin can be made by moving around 
certain points which incurs a penalty of $\{C\xi_i\}$. A high value of $C^{-}$, for instance, places a large penalty for mis-classifying instances from the negative class, and therefore 
boosts certain features from that class. As the value of $C^{-}$ increases, the model starts classifying more reviews as non-credible. In the worse case, all the reviews of an item are classified 
as non-credible, leading to the aggregated item rating being zero.

\begin{figure}
\begin{floatrow}
\hspace*{-6mm}
\ffigbox{
\includegraphics[width=\linewidth, height=5cm]{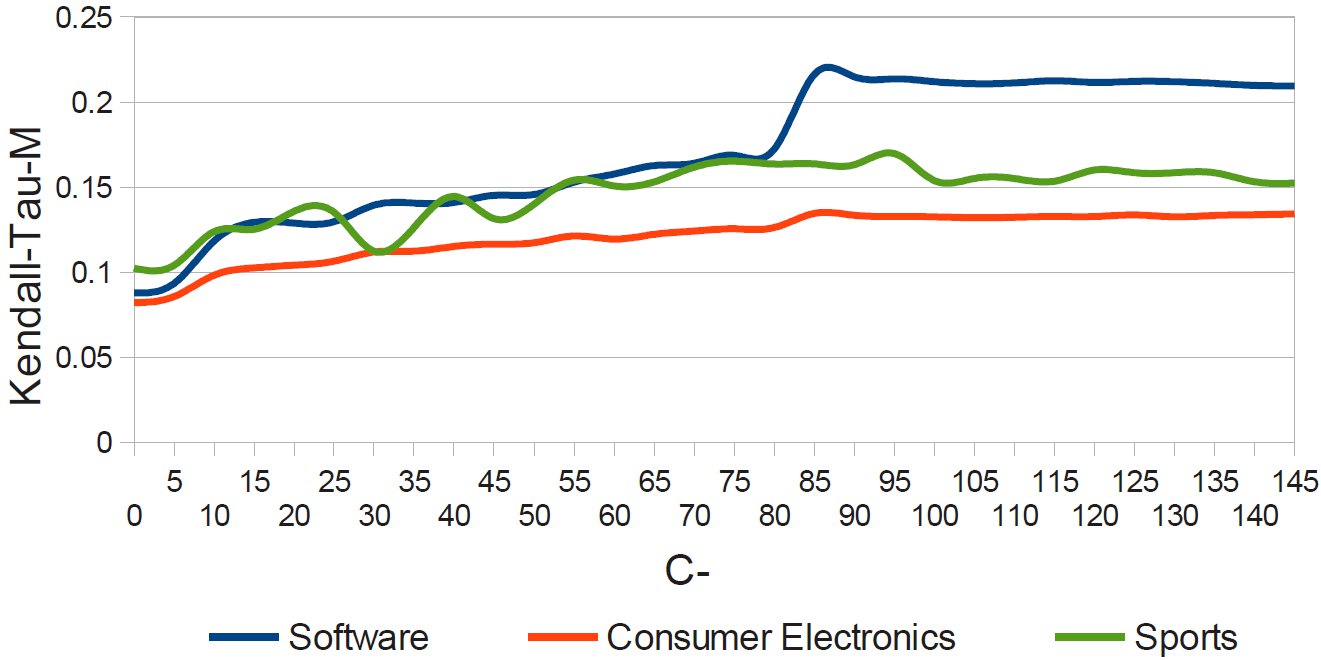}
}{
\caption{Variation of Kendall-Tau-M ($\tau_m$) on different Amazon domains with parameter $C^{-}$ variation {\scriptsize{(using model $M_\text{Yelp}$ trained in Yelp and tested in Amazon)}}.}
\label{fig:parameter}
}
\capbtabbox{
\scriptsize
\begin{tabular}{c|l}
\hline
{\bf Notation} & {\bf Description} \\ \hline
$U, D, I$ & set of users, reviews, and items resp. \\
$d, r_d$ & review text and associated rating \\
$V, f$ & unigrams and bigrams vocab. \& token types\\
$w_{ij}$ & word of token type $f_i$ in review $d_j$ \\
$I(\cdot)$ & indicator fn. for presence/absence of words \\
$z, l$ & set of facets and sentiment labels resp. \\
$K, L$ & cardinality of facets and sentiment labels \\
$\theta_d(z,l)$ & multinom. prob. distr. of facet $z$ \\
& $\quad$ with sentiment label $l$ in document $d$ \\
$\phi_{z,l}(w)$ & multinom. prob. distr. of word $w$ belonging \\
& $\quad$ to facet $z$ with sentiment label $l$ \\
$\Phi', \Phi''$ & facet-label distr. of review and item resp.\\
$\alpha, \beta, \gamma$ & Dirichlet priors \\
$\pi, \pi'$ & review rating distr. \& inferred rating distr. \\
$n(\cdot)$ & word count in reviews \\
$F^x(d_j)$ & feature vec. of review $d_j$ using lang. (x=L), \\
& $\quad$ consistency (x=T), and behavior (x=B) \\
$C^+, C^-$ & C-SVM regularization parameters \\
\hline
\end{tabular}
}
{
\caption{List of variables and notations used with corresponding description.}
\label{tab:notation}
}
\end{floatrow}
\end{figure}

We use $\tau_m$ to find the optimal value of $C^{-}$ by varying it in the interval $C^{-} \in \{0, 5, 10, 15, ... 150\}$ using a {\em validation set} from Amazon as shown in 
Figure~\ref{fig:parameter}. We observe that as $C^{-}$ increases, $\tau_m$ also increases till a certain point as more and more non-credible reviews are filtered out, after which it stabilizes.

\noindent {\em \bf Ranking SVM} --
Our previous approach uses the model $M_\text{Yelp}$ trained on Yelp, with the reference ranking (i.e., sales ranking) in Amazon being used only for evaluating the item ranking using the Kendall-Tau measure. 
As the objective is to obtain a good item ranking based on credible reviews, we can have a model $M_\text{Amazon}$ that directly optimizes for Kendall-Tau using the reference ranking as training 
labels. This allows us to use the entire feature space available in Amazon, including the explicit facet-label distribution and the full vocabulary, which could not be used earlier. The feature 
space is constructed similarly to that of Yelp.

The goal of Ranking SVM~\cite{joachims02} is to learn a ranking function which is concordant with a given ordering of items. The objective is to learn $\vec{w}$ such that $\vec{w}\cdot \vec{x_i} > \vec{w} \cdot \vec{x_j}$ for most data pairs $\{(\vec{x_i}, \vec{x_j}): y_i > y_j \in R\}$. Although the problem is known to be NP-hard, it is approximated using SVM techniques with pairwise slack variables $\xi_{i,j}$. The optimization problem is equivalent to that of classifying SVM, but now operating on {\em pairwise difference vectors} $(\vec{x_i} - \vec{x_j})$ with corresponding labels $+1 /-1$ indicating which one should be ranked ahead. We use the implementation\footnote{https://www.cs.cornell.edu/people/tj/svm\_light/svm\_rank.html} of~\cite{joachims02} that maximizes the empirical Kendall-Tau by minimizing the number of discordant pairs.

Unlike the classification task, where labels are {\em per-review}, the ranking task requires labels {\em per-item}. Consider $\langle f_{i,j, k} \rangle$ to be the feature vector for the $j^{th}$ review of an item $i$, with $k$ indexing an element of the feature vector. We aggregate these feature vectors element-wise over all the reviews on item $i$ to obtain its feature vector 
$\langle \frac{\sum_j f_{i,j, k}}{\sum_j \mathbf{1}} \rangle$.

\section{Experimental Setup}

\noindent{\bf Parameter Initialization:} The sentiment lexicon from~\cite{Hu2004} consisting of $2006$ positive and $4783$ negative polarity bearing words is used to initialize the 
review text based facet-label-word tensor $\Phi$ prior to inference. We consider the number of topics, $K=20$ for Yelp, and $K=50$ for Amazon with the review sentiment labels $L=\{+1, -1\}$ (corresponding to positive and negative rated reviews) initialized randomly. 
The symmetric Dirichlet priors are set to $\alpha = 50/K$, $\beta = 0.01$, and $\gamma = 0.1$. 

\noindent{\bf Datasets and Ground-Truth:} In this work, we consider the following datasets (refer to Table~\ref{tab:data1} and~\ref{tab:data2}) with available ground-truth information.

\begin{table}[t]
\centering
\caption{Dataset statistics for review classification. {\scriptsize{(Yelp$^{*}$ denotes balanced dataset using random sampling.)}}}
\label{tab:data1}
\begin{tabular}{p{2cm}p{4cm}p{2.5cm}p{1cm}p{1.5cm}}
\toprule
\bf{Dataset} & \bf{Non-Credible Reviews} & \bf{Credible Reviews} & \bf{Items} & \bf{Users}\\
\midrule
TripAdvisor & 800 & 800 & 20 & -\\
Yelp & 5169 & 37,500 & 273 & 24,769\\
Yelp$^{*}$ & 5169 & 5169 & 151 & 7898\\
\bottomrule
\end{tabular}
\end{table}

\begin{table}[t]
\caption{Amazon dataset statistics for item ranking, with cumulative \#items and varying \#reviews.}
\label{tab:data2}
\begin{tabular}{lrr|rrrrrrr}
\toprule
\textbf{Domain} & \multicolumn{1}{l}{\textbf{\#Users}} & \multicolumn{1}{l|}{\textbf{\#Reviews}} & \multicolumn{ 7}{c}{\textbf{\#Items with reviews per-item}} \\\midrule 
 & \multicolumn{1}{l}{} & \multicolumn{1}{l|}{} & \multicolumn{1}{l}{$\leq$5} & \multicolumn{1}{l}{$\leq$10} & \multicolumn{1}{l}{$\leq$20} & \multicolumn{1}{l}{$\leq$30} & \multicolumn{1}{l}{$\leq$40} & \multicolumn{1}{l}{$\leq$50} & \multicolumn{1}{l}{Total} \\\midrule
{\bf Consumer Electronics} & 94,664 & 121,234 & 14,797 & 16,963 & 18,350 & 18,829 & 19,053 & 19,187 & 19,518 \\ 
{\bf Software} & 21,825 & 26,767 & 3,814 & 4,354 & 4,668 & 4,767 & 4,807 & 4,828 & 4,889 \\ 
{\bf Sports} & 656 & 695 & 202 & 226 & 233 & 235 & 235 & 235 & 235 \\ 
\bottomrule
\end{tabular}
\end{table}

\noindent $\bullet$ The {\em TripAdvisor Dataset}~\cite{Ott2011,Ott2013} consists of $1600$ reviews from TripAdvisor with positive ($5$ star) and negative ($1$ star) sentiment --- comprising $20$ credible and $20$ non-credible reviews for {\em each} of $20$ most popular Chicago hotels. The authors crawled the {\em credible} reviews from online review portals like TripAdvisor; whereas the {\em non-credible} ones were generated by users in Amazon Mechanical Turk. The dataset has only the review text and sentiment label (positive/negative ratings) with corresponding hotel names, with no other information on users or items. 

\noindent $\bullet$ The {\em Yelp Dataset} consists of $37.5K$ recommended (i.e., {\em credible}) reviews, and $5K$ non-recommended (i.e., {\em non-credible}) reviews given by the Yelp filtering algorithm, on $273$ restaurants in Chicago. For each review, we gather the following information: $\langle userId, itemId, timestamp, rating, review, metadata \rangle$. The meta-data consists of some user activity information as outlined in Section~\ref{subsec:user}. 

The reviews marked as ``not recommended'' by the Yelp spam filter are considered to be the ground-truth for comparing the accuracy for credible review detection for our 
proposed model. The Yelp spam filter presumably relies on linguistic, behavioral, and social networking features~\cite{Liu2013a}.

\noindent $\bullet$ The {\em Amazon Dataset} used in~\cite{Liu2008} consists of around $149K$ reviews from nearly $117K$ users on $25K$ items from three domains, namely Consumer Electronics, Software, and Sports items. For each review, we gather the same information tuple as that from Yelp. However, the metadata in this dataset is not as rich as in Yelp, consisting only of helpfulness votes on the reviews. 

Further, there exists no explicit ground-truth characterizing the reviews as credible or deceptive in Amazon. To this end, we re-rank the items using learning to rank, implicitly filtering 
out possible deceptive reviews (based on the feature vectors), and then compare the ranking to the {\em item sales rank} considered as the pseudo ground-truth.

\noindent {\bf Comparison Baselines:} We use the following state-of-the-art baselines (given the full set of features that fit with their model) for comparison with our proposed model. 

\noindent{\em (1) Language Model Baselines:} We consider the unigram and bigram language model baselines from~\cite{Ott2011,Ott2013} that have been shown to outperform other baselines using psycholinguistic 
features, part-of-speech tags, information gain, etc. We take the best baseline from their work which is a combination of unigrams and bigrams. Our proposed model (N-gram+Facet) enriches it by using 
length normalization, presence or absence of features, latent facets, etc. 
The recently proposed {\em doc-to-vec} model based on Neural Networks, overcomes the weakness of bag-of-words models by taking the context of words into account, and learns a dense vector representation for each document~\cite{doc2vec}. We train the doc-to-vec model in our dataset as a baseline model.  
In addition, we also consider readability (ARI) and review sentiment scores~\cite{Hu2012} under the hypothesis that writing styles would be random because of diverse customer background. ARI measures the reader's ability to comprehend a text and is measured as a function of 
the total number of characters, words, and sentences present, while review sentiment tries to capture the fraction of occurrences of positive/negative sentiment words to the total 
number of such words used.

\noindent{\em (2) Activity \& Rating Baselines:} Given the tuple $\langle userId, itemId, rating, review,$ $metadata \rangle$ from the Yelp dataset, we extract all possible activity and 
rating behavioral features of users as proposed in~\cite{Liu2007,Liu2008,Liu2010,Liu2011,Liu2012,Liu2013,Liu2013a,Liu2014}. Specifically, we utilize the number of helpful feedbacks, review title 
length, review rating, use of brand names, percent of positive and negative sentiments, average rating, and rating deviation as features for classification. Further, based on the recent 
work of~\cite{Rahman2015}, we also use the user check-in and user elite status information as additional features for comparison.


\noindent {\bf Empirical Evaluations:} Our experimental setup considers the following evaluations:

\noindent {\em (1) Credible review classification:} We study the performance of the various approaches in distinguishing a {\em credible} review from a {\em non-credible} one. Since this 
forms a binary classification task, we consider a balanced dataset containing equal proportion of data from each of the two classes. On the Yelp dataset, for each item we randomly sample an equal number 
of credible and non-credible reviews (to obtain Yelp$^*$); while the TripAdvisor dataset is already balanced. Table~\ref{tab:classification} shows the $10$-fold cross validation accuracy results for the different models on the two datasets. We observe that our proposed {\em consistency and behavioral features} exhibit around $15\%$ improvement in Yelp$^{*}$ for classification accuracy over the best performing baselines (refer to Table~\ref{tab:classification}). Since the TripAdvisor dataset has {\em only} review text, the user/activity models could {\em not} be used there. The experiment could also not be performed on Amazon, as the ground-truth for credibility labels of reviews is absent.

\begin{small}
\begin{table}[t]
\centering
\caption{Credible review classification accuracy with $10$-fold cross validation. TripAdvisor dataset contains only review texts and no user/activity information.}
\label{tab:classification}
\begin{tabular}{p{3cm}p{5cm}p{2cm}p{1cm}}
\toprule
\bf{Models} & \bf{Features} & \bf{TripAdvisor} & \bf{Yelp$^{*}$} \\
\midrule
\multirow{2}{*}{\bf Deep Learning} & Doc2Vec & 69.56 & 64.84 \\
& Doc2Vec + ARI + Sentiment & 76.62 & 65.01 \\
\multirow{2}{*}{\bf Activity \& Rating} & Activity+Rating & - & 74.68 \\
& Activity+Rating+Elite+Check-in & - & 79.43 \\
\midrule
\multirow{2}{*}{\bf Language} & Unigram + Bigram& 88.37 & 73.63\\
& Consistency & 80.12 & 76.5\\
\midrule
\multirow{2}{*}{\bf Behavioral} & Activity Model$^{-}$ & - & 80.24\\
& Activity Model$^{+}$ & - & 86.35\\
\midrule
\multirow{6}{*}{\bf Aggregated} & N-gram + Consistency & {\bf 89.25} & 79.72\\
& N-gram + Activity$^{-}$ & - & 82.84\\
& N-gram + Activity$^{+}$ & - & 88.44\\
& N-gram + Consistency + Activity$^{-}$ & - & 86.58\\
& N-gram + Consistency + Activity$^{+}$ & - & {\bf 91.09}\\
& $M_\text{Yelp}$  & - & 89.87\\
\bottomrule
\end{tabular}
\end{table}
\end{small}

\noindent{\em (2) Item Ranking:} In this task we examine the effect of non-credible reviews on the ranking of items in the community. This experiment is performed {\em only} on Amazon using 
the item {\em sales rank} as ground or reference ranking, as Yelp does not provide such item rankings. The sales rank provides an indication as to how well a product is selling on 
Amazon.com and highlights the item's rank in the corresponding category\footnote{\url{www.amazon.com/gp/help/customer/display.html?nodeId=525376}}.

The baseline for the item ranking is based on the aggregated rating of all reviews on an item. The first model $M_\text{Yelp}$ (C-SVM) trained on Yelp filters out the non-credible reviews, before 
aggregating review ratings on an item. The second model $M_\text{Amazon}$ (SVM-Rank) is trained on Amazon using SVM-Rank with the reference ranking as training labels. $10$-fold cross-validation results 
are reported on the two measures of Kendall-Tau ($\tau_b$ and $\tau_m$) in Table~\ref{tab:model-ranking} with respect to the reference ranking. $\tau_b$ and $\tau_m$ for SVM-Rank are the same since there are no ties. Our first model performs substantially better than the baseline, which, in turn, is outperformed by our second model.

In order to find the effectiveness of our approach in dealing with ``long-tail'' items, we perform an additional experiment with our best performing model i.e., $M_\text{Amazon}$ (SVM-Rank). We use the model to find Kendall-Tau-M ($\tau_m$) rank correlation (with the reference ranking) of items having less than (or equal to) $5, 10, 20, 30, 40,$ and $50$ reviews in different domains in Amazon (results reported in Table~\ref{tab:long-tail} with $10$-fold cross validation). We observe that our model performs substantially well even with items having as few as {\em five} reviews, with the performance progressively getting better with more reviews per-item.

\begin{table}[t]
\caption{Kendall-Tau correlation of different models across domains.}
\label{tab:model-ranking}
\begin{tabular}{l|p{1.5cm}c|p{1.5cm}p{2.2cm}|p{2.2cm}}
\toprule
\textbf{Domain} & \multicolumn{2}{c|}{\textbf{Kendall-Tau-B ($\tau_b$)}} & \multicolumn{2}{c|}{\textbf{Kendall-Tau-M ($\tau_m$)}} & \multicolumn{1}{c}{\textbf{Kendall-Tau ($\tau_b = \tau_m$)}}\\ 
\midrule
 & \multicolumn{1}{l}{\textbf{Baseline}} & \multicolumn{1}{l|}{\textbf{$M_\text{Yelp}$ (C-SVM)}} & \multicolumn{1}{l}{\textbf{Baseline}} & \multicolumn{1}{l|}{\textbf{$M_\text{Yelp}$ (C-SVM)}} & \multicolumn{1}{l}{\textbf{$M_\text{Amazon}$ (SVM-Rank)}} \\ 
 \midrule
CE & 0.011 & 0.109 & 0.082 & 0.135 & 0.329 \\ 
Software & 0.007 & 0.184 & 0.088 & 0.216 & 0.426 \\ 
Sports & 0.021 & 0.155 & 0.102 & 0.170 & 0.325 \\ 
\bottomrule
\end{tabular}
\end{table}

\begin{table}[t]
\centering
\caption{Variation of Kendall-Tau-M ($\tau_m$) correlation with \#reviews with \textbf{$M_\text{Amazon}$ (SVM-Rank)}.}
\label{tab:long-tail}
\begin{tabular}{p{1.5cm}p{1.2cm}p{1.2cm}p{1.2cm}p{1.2cm}p{1.2cm}p{1.2cm}p{1.2cm}}
\toprule
\textbf{Domain} & \multicolumn{7}{c}{\textbf{$\tau_m$ with \#reviews per-item}} \\ 
\midrule
 & $\leq$5 & $\leq$10 & $\leq$20 & \multicolumn{1}{l}{$\leq$30} & \multicolumn{1}{l}{$\leq$40} & \multicolumn{1}{l}{$\leq$50} & \multicolumn{1}{l}{Overall} \\ \midrule
CE & 0.218 & \multicolumn{1}{l}{0.257} & \multicolumn{1}{l}{0.290} & 0.304 & 0.312 & 0.317 & 0.329 \\ 
Software & \multicolumn{1}{l}{0.353} & \multicolumn{1}{l}{0.375} & \multicolumn{1}{l}{0.401} & 0.411 & 0.417 & 0.419 & 0.426 \\ 
Sports & \multicolumn{1}{l}{0.273} & \multicolumn{1}{l}{0.324} & \multicolumn{1}{l}{0.310} & \multicolumn{1}{l}{0.325} & \multicolumn{1}{l}{0.325} & \multicolumn{1}{l}{0.325} & \multicolumn{1}{l}{0.325} \\
\bottomrule
\end{tabular}
\end{table}

\begin{table}[t]
\centering
\small
\caption{Top n-grams (by feature weights) for credibility classification.}
\label{tab:ngrams}
\begin{tabular}{p{5.7cm}p{0.2cm}p{6.3cm}}
\toprule
\textbf{Credible Reviews} & & \textbf{Non-Credible Reviews}\\\midrule
not, also, really, just, like, get, perfect, little, good, one, space, pretty, can, everything, come\_back, still, us, right, definitely, enough, much, super, free, around, delicious, no, fresh, big, favorite, lot, selection, sure, friendly, way, dish, since, huge, etc, menu, large, easy, last, room, guests, find, location, time, probably, helpful, great, now, something, two, nice, small, better, sweet, though, loved, happy, love, anything, actually, home & & dirty, mediocre, charged, customer\_service, signature\_lounge, view\_city, nice\_place, hotel\_staff, good\_service, never\_go, overpriced, several\_times, wait\_staff, signature\_room, outstanding, establishment, architecture\_foundation, will\_not, long, waste, food\_great, glamour\_closet, glamour, food\_service,  love\_place, terrible, great\_place, never,  wonderful, atmosphere, signature, bill, will\_never, good\_food, management, great\_food, money, worst, horrible, manager, service, rude \\
\bottomrule
\end{tabular}
\end{table}

\section{Discussions on Experimental Results}

{\noindent \bf Language Model:} The bigram language model performs very well (refer to Table~\ref{tab:classification}) on the TripAdvisor dataset due to the setting of the task. Workers in Amazon Mechanical Turk were tasked with writing fake reviews with the guideline of knowing all the hotel amenities in its website before writing reviews. Therefore it is quite difficult for the facet model to find contradictions or mismatch in facet descriptions. Consequently, the facet model gives marginal improvement when combined with the language model.

On the other hand, the Yelp dataset is real-world, and therefore more noisy. The bigram language model and doc-to-vec hence do not perform as good as they do in the previous dataset; and neither does the facet model in isolation. However all the components put together give significant performance improvement over the ones in isolation (around $8\%$).

Incorporation of writing style using ARI and sentiment measures improves performance of doc-to-vec in the TripAdvisor dataset, but not significantly in the real-world Yelp data.
 
Table~\ref{tab:ngrams} shows the top unigrams and bigrams contributing to the language feature space in the {\em joint model} for credibility classification --- given by the feature weights of the C-SVM. We find that credible reviews contain a mix of function and content words, balanced opinions, with the highly contributing features being mostly unigrams. Whereas, non-credible reviews contain extreme opinions, less function words and  more of sophisticated content words --- consisting of a lot of signature bigrams --- to catch the readers' attention. 

{\noindent \bf Behavioral Model:} We find the activity based model to perform the best in isolation (refer to Table~\ref{tab:classification}). Combined with language and consistency features, the joint model exhibits around $5\%$ improvement in performance. Additional meta-data like the user elite and check-in status improves the performance of activity based baselines, which are not typically available for newcomers in the community. Our model using limited information ({\em N-gram+Consistency+Activity$^{-}$}) performs better than the activity baselines using fine-grained information about items (like brand description) and user history. Incorporating additional user features ({\em Activity$^{+}$}) further boosts its performance.

{\noindent \bf Consistency Features:} In order to find the effectiveness of the facet based consistency features, we perform ablation tests (refer to Table~\ref{tab:classification}). We remove the consistency model from the aggregated model, and see significant performance degradation of $3-4\%$ for the Yelp$^{*}$ dataset. In the TripAdvisor dataset the performance reduction is less compared to Yelp due to reasons outlined before.


Table~\ref{tab:revcon} shows a snapshot of the non-credible reviews, with corresponding \\(in)consistency features in Yelp and Amazon. We see that ratings of deceptive reviews do not corroborate with the textual description, irrelevant facets influencing the rating of the target item, contradicting other users, expressing extreme opinions without explanation, depicting temporal ``burst'' in ratings, etc. In principle, these features can also be used to detect other anomalous phenomena like group-spamming (one of the principal indicators of which is temporal burst), which is out of scope of this work.

{\noindent \bf Ranking Task:} For the ranking task in Amazon (refer to Table~\ref{tab:model-ranking}), the first model $M_\text{Yelp}$ --- trained on Yelp and tested on Amazon using C-SVM --- performs much better than the baseline exploiting various consistency features. The second model $M_\text{Amazon}$ --- trained on Amazon using SVM-Rank --- outperforms the former exploiting the power of the entire feature space and domain-specific proxy labels unavailable to the former.

{\noindent \bf ``Long-Tail'' Items:} Table~\ref{tab:long-tail} shows the gradual degradation in performance of the second model $M_\text{Amazon}$ (SVM-Rank) in dealing with items with lesser number of reviews. Nevertheless, we observe it to give a substantial Kendall-Tau correlation ($\tau_m$) with the reference ranking, with as few as {\em five} reviews per-item, demonstrating the effectiveness of our model in dealing with ``long-tail'' items.

\begin{table}[t]
\centering
\caption{\small Snapshot of non-credible reviews (reproduced verbatim) with inconsistencies.}
\label{tab:revcon}
\begin{tabular}{p{2.4cm}|p{4.5cm}|p{5.5cm}}
\toprule
{\bf Inconsistency Features} & {\bf Yelp} Review \& [Rating] & {\bf Amazon} Review \& [Rating] \\
\midrule
{\bf user review -- rating} {\em (promotion/demotion)}:  & 
\underline{never been inside James.} \underline{never checked in.} \underline{never visited bar.} yet, one of my favorite hotels in Chicago. James has dog friendly area. my dog loves it there. [5] & 
Excellant product-alarm zone, technical support is almost non-existent because of this \underline{i will look to another product.} \underline{this is unacceptible}. [4]\\
{\bf user review -- facet description} {\em (irrelevant)}: &
you will learn that they are actually \underline{EVANGELICAL CHRISTIANS} working to proselytize the coffee farmers they buy from. [2] &
DO NOT BUY THIS. I used turbo tax since 2003, it never let me down until now. I can't file because Turbo Tax doesn't have software updates from the IRS \underline{``because of Hurricane Katrina''}. [1] \\
{\bf user review -- item description} {\em (deviation from community)}: & 
{internet is charged} in a $300$ dollar hotel! [3] &
The book Amazon offers is a joke! All it provides is the forward which is not written by Kalanithi. I don't have any \underline{sample of} \underline{HIS writing} to know if it appeals. [1] \\
{\bf extreme user rating}: &
{GREAT!!!i give 5 stars!!!Keep it up.} [5] &
GREAT. This camera takes pictures. [1]\\
\multirow{4}{*}{\bf temporal bursts\footnotemark:} &
\multicolumn{2}{|l}{Dan's apartment was beautiful and a great downtown location... \underline{(3/14/2012)} [5]} \\
& \multicolumn{2}{|l}{I highly recommend working with Dan and NSRA... \underline{(3/14/2012)} [5]} \\
& \multicolumn{2}{|l}{Dan is super friendly, demonstrating that he was confident... \underline{(3/14/2012)} [5]} \\
& \multicolumn{2}{|l}{my condo listing with no activity, Dan really stepped in... \underline{(4/18/2012)} [5]} \\
\bottomrule
\end{tabular}
\end{table}
\footnotetext{{\scriptsize these reviews have also been flagged by the Yelp Spam Filter as not-recommended (i.e., non-credible)}}


\section{Conclusions}

We present a novel consistency model using limited information for detecting non-credible reviews which is shown to outperform state-of-the-art baselines. Our approach overcomes the limitation of existing works that make use of fine-grained information which are not available for ``long-tail'' items or newcomers in the community. Most importantly prior methods are not designed to {\em explain} why the detected review should be non-credible. In contrast, we make use of different {\em consistency features} from {\em latent facet model} derived from user text and ratings that can explain the assessments by our method. We develop multiple models for domain transfer and adaptation, where our model performs very well in the ranking tasks involving ``long-tail'' items, with as few as {\em five} reviews per-item.


\bibliographystyle{splncs03}
\bibliography{ecml16}

\end{document}